\documentclass[10pt,twocolumn,letterpaper]{article}

\usepackage{cvpr}
\usepackage{times}
\usepackage{epsfig}
\usepackage{graphicx}
\usepackage{amsmath}
\usepackage{amssymb}
\usepackage{algorithm}
\usepackage{algorithmic}
\usepackage{listings}
\usepackage{xcolor}
\usepackage{booktabs}
\usepackage{comment}
\usepackage{balance}

\usepackage{appendix}
\usepackage{xcolor}

\definecolor{codegreen}{rgb}{0,0.6,0}
\definecolor{codegray}{rgb}{0.5,0.5,0.5}
\definecolor{codepurple}{rgb}{0.58,0,0.82}
\definecolor{backcolour}{rgb}{0.95,0.95,0.92}

\lstdefinestyle{mystyle}{
    frame=single,
    commentstyle=\color{codegreen},
    keywordstyle=\color{magenta},
    numberstyle=\tiny\color{codegray},
    stringstyle=\color{codepurple},
    basicstyle=\ttfamily\footnotesize,
    breakatwhitespace=false,         
    breaklines=true,                 
    captionpos=b,                    
    keepspaces=true,                 
    numbersep=5pt,                  
    showspaces=false,                
    showstringspaces=false,
    showtabs=false,                  
    tabsize=2
}

\usepackage[pagebackref=true]{hyperref}

\begin{document}

\title{A Self-Supervised Learning Pipeline for Demographically Fair Facial Attribute Classification}


\author{Sreeraj Ramachandran\\
Wichita State University\\
Wichita, Kansas, USA\\
{\tt\small sxramachandran2@shockers.wichita.edu}
\and
Ajita Rattani\\
University of North Texas \\
Denton, Texas, USA\\
{\tt\small ajita.rattani@unt.edu}
}

\maketitle
\thispagestyle{empty}

\begin{abstract}
Published research highlights the presence of demographic bias in automated facial attribute classification. The proposed bias mitigation techniques are mostly based on supervised learning, which requires a large amount of labeled training data for generalizability and scalability. However, labeled data is limited, requires laborious annotation, poses privacy risks, and can perpetuate human bias. In contrast, self-supervised learning (SSL) capitalizes on freely available unlabeled data, rendering trained models more scalable and generalizable. However, these label-free SSL models may also introduce biases by sampling false negative pairs, especially at low-data regimes (\(<200\)K images) under low compute settings. Further, SSL-based models may suffer from performance degradation due to a lack of quality assurance of the unlabeled data sourced from the web. 
This paper proposes a fully self-supervised pipeline for demographically fair facial attribute classifiers. Leveraging completely unlabeled data pseudolabeled via pre-trained encoders, diverse data curation techniques, and meta-learning-based weighted contrastive learning, our method significantly outperforms existing SSL approaches proposed for downstream image classification tasks. Extensive evaluations on the FairFace and CelebA datasets demonstrate the efficacy of our pipeline in obtaining fair performance over existing baselines. Thus, setting a new benchmark for SSL in the fairness of facial attribute classification.

\end{abstract}
\vspace{-0.2cm}
\section{Introduction}

Automated facial analysis-based algorithms encompass face detection, face recognition, and facial attribute classification (such as gender-, race-, age classification, and BMI prediction)~\cite{facerecsurvey,facedetectionsurvey,celeba,SiddiquiRRH22}. Numerous existing studies~\cite{grother,KlareBKBJ12,Best-RowdenJ18,Vera-RodriguezB19,AlbieroSVZKB20,KrishnanAR20} confirmed the performance disparities of facial attribute classifiers between demographic groups. 
Most of these bias mitigation techniques are predominantly \textbf{supervised learning techniques}~\cite{ZhangLM18,ChuangM21,ramachandran2023leveraging,DasDB18,LinKJ22,KRISHNAN2023104793}, that aim to introduce fairness constraints during model training using demographically annotated training data. However, sensitive demographic data is often laboriously annotated through large-scale human efforts, raising privacy issues and General Data Protection Regulation~(GDPR)~\cite{gdpr} related legal constraints. Further, labeled data are often scarce in real-world applications, leading to limited generalizability and scalability (due to higher annotation costs) of the trained models.
Further, there is a potential perpetuation of human biases from the data labeling process~\cite{Sun2020-am}. Further, these supervised bias mitigation techniques are often Pareto inefficient~\cite{ZietlowLBKLS022} which means fairness is often obtained at the cost of reduced overall performance.

This prompts a shift towards \textbf{self-supervised learning} (SSL), which can take advantage of the abundance of unlabeled data using contrastive learning~\cite{contrastive,simclr, clip,laion} for model training. Contrastive learning aims to create dense representations that exhibit low intra-class variance and high inter-class variance even without explicit labels.
Thus, there is no requirement for annotated labels and sensitive attributes allowing for highly scalable training methodologies, reducing manual effort, and bypassing privacy concerns related to sensitive information. Readers are referred to Appendix A for background mathematical details on contrastive loss.

However, this label-free approach of SSL can also introduce bias by inadvertently sampling \textit{false negative pairs} i.e., in a single batch, it might inadvertently pair two images of the same class (unknown) as negative pairs~\cite{contrastive,simclr,Khosla2020,clip}. 
This issue is particularly relevant in \textbf{low-compute settings}, where training on massive datasets is impractical due to time and resource constraints. Therefore, this study focuses on a low-data regime (\textless200K images) to investigate the effectiveness of SSL under such constraints, as the limited number of total data points increases the chances of sampling false negative pairs, thus amplifying biases on the learned representation. 
Further, these SSL methods may also suffer from the issue of lack of \textit{quality assurance}, a result of extensive web-scraped datasets from the wild.

We \textbf{aim} to propose a fully SSL-based pipeline for fair representation learning that addresses the aforementioned limitations of current supervised and SSL approaches. 

To this front, our method integrates a multitude of techniques to devise an end-to-end fully self-supervised pipeline consisting of (a) \textbf{data curation}, and (b) \textbf{training pipeline}. 
Specifically, our data curation pipeline intends to enhance the diversity, reduce redundancy, and enhance quality of the unlabeled training data by employing deduplication and self-supervised image retrieval strategies. 
This is crucial for training unbiased models that allow us to operate in a low-data regime SSL-based training (\textless200K images) under low compute settings. 

Further, addressing the limitation of widely adopted classical contrastive learning (refer Appendix A for mathematical details) in SSL approaches,  we notably adapt the supervised contrastive loss~(SupCon)~\cite{Khosla2020}, which usually requires supervisory signals through labeled samples~\cite{fscl}, to operate in a self-supervised mode using pseudo-labeled samples in the training pipeline, obtained using state-of-the-art pretrained encoders, such as CLIP~\cite{clip} and DINO~\cite{dino}, in a zero-shot setting~\cite{clip} (refer Appendix B for details). 

Further, we also adopted a weighted meta-learning-based strategy during the training stage to assign weights to individual samples based on gradient signals obtained from the pseudo-labeled validation set to improve both performance and fairness. 

In summary, the main \textbf{contributions} of this paper are as follows
\begin{itemize}
\item \textbf{Data Curation Pipeline}: Develops a scalable pipeline that leverages curated datasets like FairFace~\cite{fairface} to produce high-quality unlabeled datasets automatically, enhancing data diversity and quality without manual labor.

\item \textbf{SupCon Loss with Pseudo Labels}: Adapts and employs Supervised Contrastive Loss with pseudo labels generated via zero-shot techniques, offering a robust alternative to standard contrastive loss and improving model learning efficiency on unlabeled data.

\item \textbf{Meta-Weight Learning with Automated Labeling}: Introduces a meta-weight learning algorithm that requires no manually labeled samples, using automated pseudo-labeling to optimize fairness as well as accuracy, a notable advancement over traditional methods that rely on labeled data.

\item \textbf{Automated, Scalable SSL Pipeline for Fair Representation Learning}: Combines advanced data curation, SupCon Loss with pseudo labels, and meta-weight learning into a fully automated, scalable pipeline for fair representation learning in SSL, thus setting a new benchmark in the field of FairAI.

\end{itemize}

Thus our contributions are manifold and signify a leap forward in fair representation learning for facial attribute classifiers using SSL. 

\section{Related Work}

\textbf{Fairness in Facial Attribute Classification}:  Research has consistently shown that facial-attribute classification algorithms exhibit significant biases, affecting accuracy across different gender-racial groups and exacerbating societal inequalities~\cite{face_bias_survey,gendershade,Muthukumar19,KrishnanAR20,BarlasKGKO20}. 
In response, a variety of supervised learning-based mitigation strategies 
have been developed~\cite{MajumdarSV21,face_bias_survey,ramachandran2022deep,KRISHNAN2023104793} to mitigate bias in facial attribute classifiers.
For example, ~\cite{gbr} proposed learning fair representations by reversing gradients of the classification loss for sensitive attributes through a gradient reversal layer. ~\cite{fnl} extended this approach by minimizing the mutual information between the representation and sensitive attribute labels to further reduce their correlations. In FD-VAE~\cite{fdvae}, latent code for sensitive attributes capture information related to both target and sensitive attributes, which is then discarded to ensure fairness. These mitigation techniques are often Pareto inefficient~\cite{ZietlowLBKLS022,ValdiviaSC21,MartinezBS20} and lack scalable and generalizability~\cite{ZietlowLBKLS022,ramachandran2022deep,ZhangLM18,ChuangM21}.

\textbf{SSL Approaches}: Self-supervised methods such as SimCLR~\cite{simclr} maximize and minimize the similarity between augmented views of the same image and different images using a contrastive loss, a large batch size, and a projection head. Whereas BYOL~\cite{byol} learns representations by predicting the target network's representation of a transformed view of the same image using an online network without using negative pairs. Barlow Twins~\cite{barlow} learns representations by minimizing the redundancy between the output representation dimensions, enforcing off-diagonal elements of the cross-correlation matrix between distorted views to be zero while preserving diagonal elements. VicReg~\cite{vicreg} learns representations by regularizing the variance and covariance of the embeddings along each dimension individually. Finally, the DINO~\cite{dino} method learns representations by matching the output of a student network to a teacher network, where the teacher network is updated using an exponential moving average of the student network's parameters. DINOV2~\cite{dinov2} improves on it by introducing data curation methods to improve generalization by finding images similar to curated data sets using embedding similarities. 
Finally, a method proposed by ~\cite{abs-2403-02138} uses a facial-region-aware SSL technique for a large-scale facial analysis model trained on $3.3M$ images. As our work focuses on low data regimes and compact models, a direct comparison with~\cite{abs-2403-02138} is beyond the scope of this study.

\section{Proposed Method}
In this section, we will discuss our data curation and training pipeline in detail.
\begin{figure*}[!hbtp]
    \centering
    \includegraphics[width=0.89\textwidth]{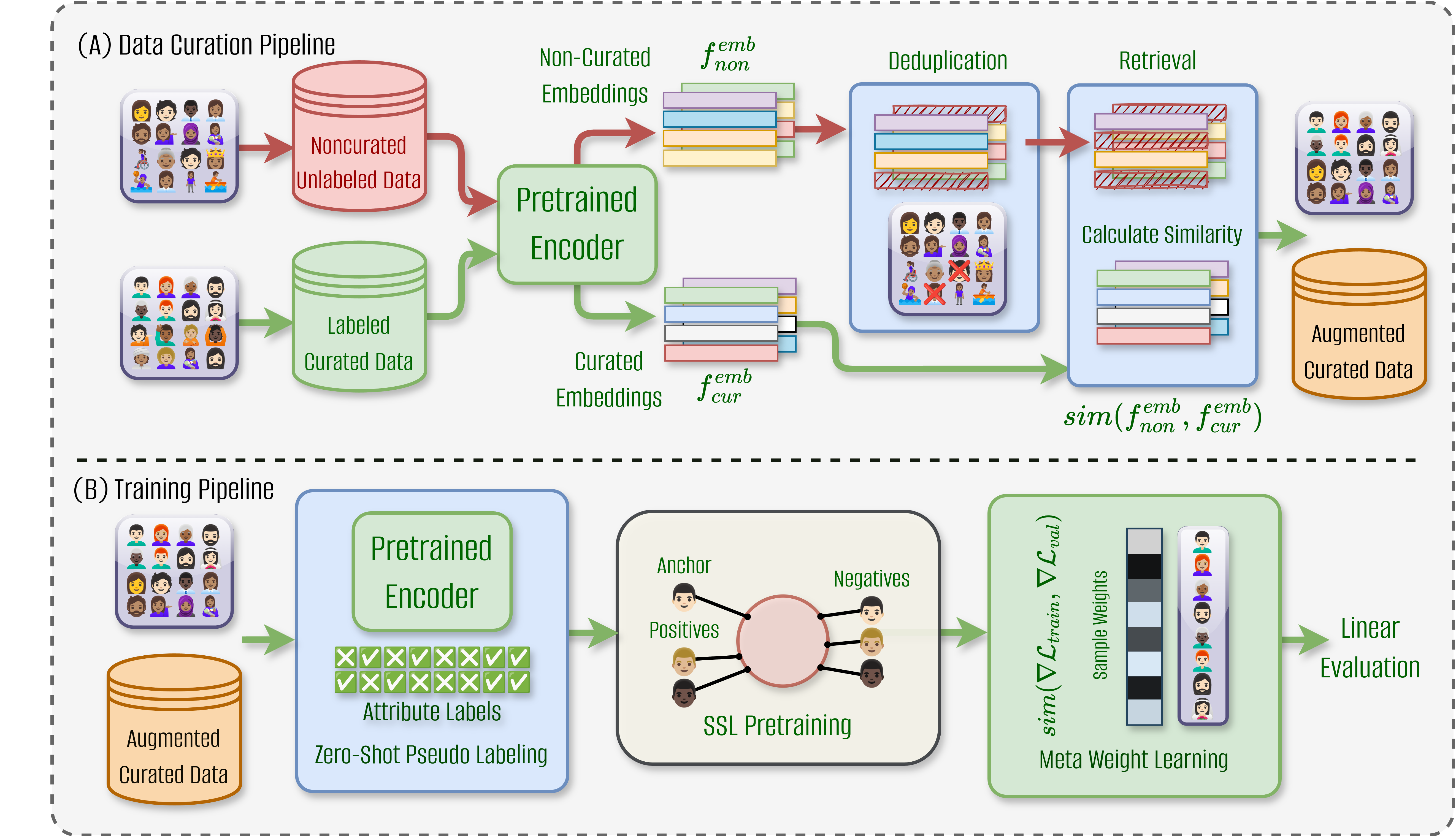}
    \caption{Overview of the proposed SSL framework. A) Data Curation Pipeline: Embeddings for unlabeled data are generated using a pre-trained encoder and used to deduplicate the data. The similarity between curated and noncurated embeddings is used to retrieve similar samples from the noncurated set to generate the final augmented curated set. (B) Training Pipeline: The encoder is trained using the augmented curated data and attribute labels obtained using zero-shot pseudo-labeled. SSL pretraining is performed using the SupCon loss, followed by weighted meta-learning to improve performance. Finally, a linear evaluation (probing) is conducted for the quality assessment of the learned embeddings.} 
    \label{fig:ssl_teaser}
\end{figure*}

\subsection{Data Curation Pipeline}
\label{subsec:data_curation}
To train a self-supervised model that can reach the level of performance of a model trained using supervised methods typically requires a substantial amount of unlabeled raw data often scraped from the Web. For example, the state-of-the-art CLIP model~\cite{clip} was trained on a massive dataset of $400$ million images to achieve the same accuracy in linear-probe testing as a comparatively smaller model (EfficientNet) trained on only approximately $1.2$ million images from the ImageNet dataset. Given the intractability of such large-scale training in a low-compute setting, our approach diverges from this standard by concentrating on smaller models and compact datasets, i.e., the size of carefully curated datasets is $\leq 200,000$ images. 
Given this scale, the selection and curation of our training data demand exceptional quality that has a diverse and balanced distribution. In order to meet these requirements, we have applied a new data curationpipeline taking cues from the innovative approaches proposed in DinoV2 paper~\cite{dinov2} and applying an implicit fairness constraint by reproducing the data distribution of existing fair and balanced datasets such as FairFace~\cite{fairface}. An overview of the data curation pipeline is illustrated in Fig.~\ref{fig:ssl_teaser}.

\noindent \textbf{Data Sources}: In the first stage of the data curation pipeline, we gather a large collection of unlabeled datasets, aiming to cover a wide variety of face images.  We refer to this collection as our \textit{unlabeled/noncurated} dataset. It primarily includes a randomly sampled subset from two prominent sources: the LAION Face Dataset~\cite{laion} and the BUPT Face Dataset~\cite{bupt}. While these datasets offer a wealth of images, varying in quality and resolution ($\sim 50M$), they lack the specific facial attribute labels needed for our downstream task of facial attribute classification.

Whereas a dataset like FairFace~\cite{fairface}($\sim 100K$) was curated and \textit{manually annotated} with the objective of a balanced distribution to be used in fair attribute classification training and evaluation. In our approach, we want to emulate its fair and diverse data distribution, implicitly, through automated retrieval methods. To do this, we first process both our uncurated collection and the FairFace dataset(\textit{retrieval set}) through a pre-trained ViT-L/16 vision encoder of the CLIP model. This step produces image embeddings, which are essentially numerical representations that capture the essence of each image.

\noindent \textbf{Deduplication and Retrieval}: In the second stage, we use the generated embeddings for data filtering. To ensure we keep only unique images, we first run a deduplication step on our noncurated dataset using cosine similarity on the image embeddings to identify and remove duplicates. With a deduplicated set, we can now perform a similarity search to find and retrieve images that are close matches to the good-quality, diverse examples found in the FairFace dataset (retrieval set). For each image in the retrieval set, we retrieve \(M\) similar embeddings from the noncurated set. A relatively small value for \(M (<5)\) was found to be sufficient in retrieving good-quality images that closely mirrored the distribution of the retrieval set and the rest was discarded. And this also speeds up the querying process. In order to carry out this search quickly and efficiently, we employed the Faiss~\cite{faiss} package from Meta, which provides KNN search capabilities. This library helps us sift through the vast array of image embeddings using GPU acceleration to find the best matches. At this point, we have a dataset of approximately $500K$ images.

\noindent \textbf{Image Quality Filtering}: In the final stage, we have the option to filter our images further with a quality assessment model. This optional step involves utilizing a non-reference-based facial image quality assessment tool, CR-FIQA proposed in~\cite{Boutros_2023_CVPR} to generate quality scores for the filtered set. We may then subsequently filter out lower-quality images by setting up an appropriate threshold based on empirical evidence. After this step, our final dataset size is $\sim 200K$ images, similar to that of the CelebA~\cite{celeba} dataset.

The result of this \textit{carefully designed} data curation pipeline is an \textit{augmented curated} dataset — a combination of the noncurated images that have passed our quality and similarity checks, along with the initial curated set (FairFace). This filtered dataset delivers the necessary diversity and balance for our facial attribute classification tasks, automating what traditionally required extensive manual labor.

\subsection{Self-supervised Training Pipeline}
\label{sec:ssl}
In this section, we outline how the \textit{augmented-curated} dataset is integrated into a self-supervised training pipeline to operate in a low-data regime. An overview of the full training process is illustrated in Fig.~\ref{fig:ssl_teaser}.
\vspace{-0.2cm}

\subsubsection{Pseudolabeling using zero-shot techniques}
\label{subsec:pseudolabel}
In this stage, we use a pretrained CLIP model to create pseudo labels for our augmented-curated dataset. We specifically follow the zero-shot classification method outlined in the CLIP paper~\cite{clip}. Readers are referred to Appendix B for more details. We also note that the choice of encoder is crucial here, as biases from the encoder may persist. We hypothesize that strategically selecting a fair encoder trained on multiple tasks across various domains would yield better results than our arbitrary choice of CLIP.

This decision to generate pseudo labels stems from our intention to employ Supervised Contrastive Loss (\textit{SupCon} Loss)~\cite{Khosla2020} instead of standard contrastive loss (refer Appendix A for mathematical details). The SupCon loss is specifically adapted to leverage the structured information that pseudo-labels provide and to eliminate the issue of false negative pairs, a known challenge of label-independent contrastive learning for SSL models. Given pseudo labels, naturally arises the question of training using the standard cross-entropy loss used for classification. However, pseudo-labels often imply noisy and low-quality labels. In such scenarios, the SupCon loss proves to be more robust and can handle the inherent uncertainty of pseudo labels because it concentrates on relative rather than absolute label assignment. Therefore, the model can generalize better and produce more adaptable representations for downstream classification tasks~\cite{Khosla2020}.
\vspace{-0.2cm}
\subsubsection{Supervised Contrastive Learning}
\label{sec:supcon}
Moving to supervised scenarios where labels are available whether manually labeled or pseudo-labeled, standard contrastive loss (readers are referred to Appendix A for more information on contrastive loss.) needs adjustment to account for labeled data where multiple samples can be of the same class. To handle this, SupCon Loss extends the standard contrastive loss by modifying it to include supervisory signals. 
\begin{equation}
\begin{aligned}
\mathcal{L}^{supcon} &= \sum_{i \in I} \frac{-1}{|P(i)|} \sum_{p \in P(i)} \log \left( \frac{\exp(z_i \cdot z_p / \tau)}{\sum_{a \in A(i)} \exp(z_i \cdot z_a / \tau)} \right)
\end{aligned}
\label{eq:supcon}
\end{equation}
Here, \(P(i)\) refers to the set of positive indices linked with \(i\)th image, sharing the same label in the multiviewed batch (set of augmented samples), but excluding \(i\) itself. \(z_i\) is the output feature embedding of the augmented views, \(\tau\) is a positive scalar known as the temperature parameter used to scale the logits, and \(A(i)\) is the set of all images except the \(i\)th image.

Thus, overall, in the supervised contrastive learning stage, we first create two augmented versions for every batch of pseudo-labeled input data. These augmented versions are fed into an encoder network, producing embeddings of $512$ dimensions. These embeddings are then normalized to the unit hypersphere. During the training process, these embeddings are further processed through a projection network, normally a 3-layer MLP network. It's important to note that this projection network is not used after the pretraining stage. The supervised contrastive loss is calculated based on the output feature embeddings from this projection network using eq.~\ref{eq:supcon}. 
Finally, we add a linear classifier on top trained using cross-entropy loss to achieve the final classification performance, a process called \textit{linear probing}.

\noindent\textbf{What attributes to use?} The use of SupCon loss typically ties a model with the specific task due to its dependency on attribute labels (here pseudo-labels). To avoid limiting our model to a specific attribute classification task, we use all 40 attributes (Young, Beard, Attractive, etc.) similar to those annotated for the CelebA~\cite{celeba} dataset, in the pseudo-label generation phase discussed in section~\ref{subsec:pseudolabel} (zero-shot, refer Appendix B for details) to generate 40 different pseudo-labels for each sample in the augmented curated dataset.  
This results in a model that could be generalized across a range of facial attributes. When calculating the loss, we determined the SupCon loss for each attribute individually and then computed the mean of these losses to obtain the overall loss measure. 

\subsection{Meta-Weighted SupCon with Fairness Constraints}
\label{sec:meta}
\subsubsection{Fairness Constraint}
Until this stage, our training stage does not explicitly impose a fairness constraint (other than the implicit fairness constraint by replicating Fairface distribution during data curation). Since we aim to develop a model that generates fair embeddings without access to explicit labels, applying any fairness constraint that involves explicit sensitive attribute labeling naturally becomes an issue. 

To this front, we adopt the label-free Min-Max fairness constraint, drawing inspiration from ~\cite{Hashimoto2018, ssl}. Specifically, our objective is to minimize the average of the top-\(k\) losses which can be represented equivalently as:
\begin{align}
\label{eq:supcon_fair}
\begin{split}
\mathcal{L}^{supcon}(k, \theta) &= \frac{1}{k} \sum_{i=1}^{2N} \max\{\mathcal{L}^{supcon}(\tilde{x}_i; \theta) - \lambda(k, \theta), 0\} \\ & \quad + \lambda(k, \theta),
\end{split}
\end{align}
In this equation, \( \lambda(k, \theta)\) signifies the \(k\)-th highest contrastive loss within the set of all contrastive losses in a batch and \(\mathcal{L}^{supcon}(\tilde{x}_i; \theta)\) is the SupCon loss from eq.~\ref{eq:supcon}. Readers are referred to Appendix C for mathematical background on the Min-Max fairness constraint.

\textbf{Challenge}: Despite its advantages, SupCon loss still faces challenges when dealing with noisy labels and potential class imbalance. Specifically, the inherent noise in our pseudo-labeling process, coupled with the possibility of an imbalanced dataset, can introduce bias during training. Specifically, if certain classes are underrepresented, the model may not learn equally effective representations for all classes, impacting the overall fairness and performance of the downstream classification task.

\textbf{Solution through proxy}: The work by ~\cite{ssl} addresses this issue by applying the fairness constraints on a small but manually labeled validation set first. Then the corresponding validation loss is used as a proxy for the training objective. Our method further improves on this by eliminating the need for the manually annotated validation set, making it fully compatible with the self-supervised learning pipeline. Thus, instead of a small, labeled validation set, we sample a small subset of images, which have been pseudo-labeled with high confidence, from our augmented curated dataset. 

The validation loss therefore becomes,
\begin{equation}
\label{eq:m_loss}
\begin{aligned}
\lambda^{val}(k, \theta, \omega) = \biggl[ \frac{1}{k} \sum_{j=1}^{M} & \left[ \mathcal{L}_{cls} \left(g_\omega \left(f_\theta(\hat{x}_j)\right), \hat{y}_j\right) \right. \\
& \left. - \lambda^{val}(k, \theta, \omega)\right]_+ \\
& + \lambda^{val}(k, \theta, \omega) \biggr],
\end{aligned}
\end{equation}
where instead of \(\lambda^{val}(k, \theta\) from eq.~\ref{eq:supcon_fair} we have \(\lambda^{val}(k, \theta, \omega)\) which denotes the \(k\)-th largest classification loss within the subset of the curated set with high-confidence pseudo labels \(\{\mathcal{L}_{cls}(g_\omega(f_\theta(\hat{x}_j)), \hat{y}_j)\}_{j=1}^{M}\), \(g_\omega\) is the additional linear layer used for computing the validation loss and \(f_\theta\) is the base model.

\subsubsection{Meta Weight Learning}
To use the obtained validation loss as a proxy during the training, we follow a similar approach to that of the works by ~\cite{ssl} and reweighting techniques ~\cite{reweighting}. These methods work by dynamically adjusting the influence of individual training samples to align more closely with the desired objective; in our case, validation loss with fairness constraint was applied. Given the defined validation loss \(\lambda^{val}(k, \theta, \omega)\) in eq.\ref{eq:m_loss}, our goal is to iteratively refine the weights of our training samples in a manner that optimally balances the competing demands of maintaining a low contrastive loss while adhering to the fairness constraints inferred from the pseudo-labeled validation subset. This balance is crucial for ensuring that our model learns effective and generalized representations and does so in a manner that respects the fairness constraint.
\vspace{-0.13 cm}

\textbf{Weight Update Mechanism}: To achieve this, we introduce a weight update mechanism that weights individual training samples dynamically. Specifically, each training sample, \(\hat{x}_I \) is assigned a dynamic weight \(w_i\) that reflects its current contribution towards meeting the combined objectives of contrastive loss as well as the validation loss. The weight of the training sample is updated based on the gradient similarity between training loss and validation loss~\cite{reweighting}. Specifically, for each training sample, we calculate the gradient of the supervised contrastive loss with respect to the model parameters as \(\nabla_{\theta} \mathcal{L}^{supcon}(f_{\theta}(\hat{x}_i), \hat{y}_i)\), and adjust its weight \(w_i\) according to the similarity of this gradient to that of the validation loss and normalizes it, reinforcing samples that guide the model towards fairness. Finally, the weighted SupCon loss is recalculated using the updated weights, and backpropagation is applied. These gradient values can be estimated using forward-mode automatic differentiation libraries such as PyTorch using multiple forward and backward passes~(refer Algorithm in Appendix D).

\textbf{Training Challenges and Mitigation Strategy}: The meta-weighting strategy introduced above also comes with a caveat of increased computational complexity. As evident from the algorithm described in Appendix D, it requires two forward and three backward passes in each training iteration, a significant increase from the standard single forward and backward passes. This modification nearly triples the training time and demands additional memory to store those intermediate states, presenting notable challenges in computational efficiency and memory resource allocation.

To address this challenge, we apply our strategy selectively 
during the later stages of training. For an initial portion of the training process (determined by hyperparameter, $k$), we employ the standard SupCon loss in eq.~\ref{eq:supcon}. In the later half of the training, we freeze certain layers of the network and apply the meta-weighting strategy only to the remaining actively trained layers. This staged training substantially enhances the efficiency by reducing the computational load during the initial training phase. Furthermore, it ensures that the network benefits from the fine-tuning capabilities of the meta-weighting scheme in the critical later stages. For example, with $k$ equal to $0.7$, i.e., $210$ out of $300$ epochs trained using SupCon and rest applying the meta-weighting strategy on the unfrozen layers, we obtain $\sim3\times$ faster training speed up.

\section{Experiments and Results}
\subsection{Dataset}
\label{subsec:datasets}
For our experiments, we used the FairFace~\cite{fairface} as our curated dataset.  For evaluation purposes, we used FairFace for intra-dataset and CelebA~\cite{celeba} for cross-dataset evaluation. For the non-curated dataset, we used both the LAION-FACE~\cite{laion} dataset as well as the BUPT-GlobalFace Dataset~\cite{bupt}, which are not annotated with any task-specific attribute labels. More information about the datasets used is available in Appendix E.

\subsection{Experimental Configuration and Metrics}
\label{sec:experiment_arch}
\textbf{Model Architecture}: For all our experiments, we used a ResNet-18 architecture as the backbone encoder. Although other architectures were also considered, we used this architecture because ours is a low-data regime approach, and a small model was deemed more appropriate. For the contrastive training, the Resnet-18 encoder takes the augmented views as input. It is then followed by the projection network, which is a 3-layer multi-layer perception~(MLP) with a hidden layer size of $2048$ and output dimension of $512$. The projection network produces the final representation used for the corresponding final loss calculation. 
After the aforementioned contrastive pre-training, the Projection Network is discarded, making the inference-time model architecture identical to a standard cross-entropy model utilizing the same encoder. 
During the linear probe evaluation phase, we kept the encoder frozen and added a linear layer at the end to be trained via Logistic Regression for each attribute, following a typical self-supervised learning protocol.

\textbf{Training hyperparameters and SSL Baselines}: For training all our models, we used two Nvidia A$6000$ GPUs as our computational infrastructure. The training was done in batches of $256$ over $300$ epochs. We used a learning rate scheduler with cosine decay with warmup, a base learning rate of $1e-4$, and a weight decay of $5e-4$ using AdamW~\cite{adamw} optimizer. We applied the same data augmentation strategy from the SimCLR~\cite{simclr} paper, i.e., we sequentially apply 3 simple augmentations: random cropping and resize, random color distortions, and random Gaussian blur. We then benchmarked our method against several SSL approaches namely, SimCLR~\cite{simclr}, BYOL~\cite{byol}, Barlow Twins~\cite{barlow}, VicReg~\cite{vicreg}, and DINO~\cite{dino}. We also compared our method against supervised bias mitigation techniques namely, GRL~\cite{gbr}, LNL~\cite{fnl}, and FD-VAE~\cite{fdvae}, whenever feasible.

We report the results from our proposed approach under two configurations, (1) \textbf{Ours}: uses the data curation pipeline, pseudo labeling along with the SupCon loss modification (See Section~\ref{sec:ssl}) and (2) \textbf{Ours Weighted}: uses, in addition to that, fairness constraints and meta-weighted learning from Section~\ref{sec:meta}.

Additionally, we compared our proposed approach to the same backbone ResNet-18 model trained using supervised cross-entropy loss. However, note that comparing our SSL approach trained from scratch to the supervised fine-tuned (ImageNet weights) model may not be entirely fair. 
Therefore this comparison serves primarily as a reference to gauge the performance potential of our SSL pipeline, rather than as a direct, fair comparison between the methods.

\begin{table*}[ht]
\centering
\scalebox{0.9}{
\begin{tabular}{@{}lcccccc@{}}
\toprule
\textbf{Config} & \textbf{Avg. Acc \textuparrow} & \textbf{STD \textdownarrow} & \textbf{SeR \textuparrow} & \textbf{EOD \textdownarrow} & \textbf{Min Grp Acc \textuparrow} & \textbf{Max Grp Acc \textuparrow} \\ \midrule \midrule
SimCLR~\cite{simclr} & 85.69 & 6.33 & 75.86 & 10.38 & 70.54 & 92.99 \\
DINO~\cite{dino} & 85.99 & 5.64 & 74.13 & 10.12 & 68.76 & 92.76 \\
BYOL~\cite{byol} & 86.07 & 5.31 & 77.47 & 9.98 & 71.47 & 92.25 \\
VicReg~\cite{vicreg} & 84.36 & 6.74 & 72.60 & 9.99 & 66.97 & 92.25 \\
BarlowTwins~\cite{barlow} &    84.22 & 6.60 & 73.70 & 9.64 & 67.90 & 92.13 \\
\midrule
Supervised (From Scratch) & 90.29 & 2.85 & 84.12 & 7.20 & 80.20 & 95.33 \\
Supervised (Finetuned Imagenet) & 94.15 & 2.25 & 92.25 & 3.24 & 88.67 & 96.25 \\
\midrule
Ours  & 91.09 & \textbf{2.59} & \textbf{88.50} & \textbf{6.24} & \textbf{84.15} & 95.08 \\
Ours-Weighted & \textbf{91.37 }& 2.91 & 86.42 & 7.05 & 82.69 & \textbf{95.69} \\
\bottomrule
\end{tabular}
}
\caption{Linear Probe Evaluation Results on Gender Classification Task on FairFace. All the models, including the supervised model, are based on the ResNet-18 backbone. \textbf{Ours}: Data curation pipeline + Pseudo Label + SupCon Loss. \textbf{Ours-Weighted}: Ours + Fairness Constraint + Meta Learning.} 
\label{tab:results_fairface}
\end{table*}

\textbf{Metrics Evaluated}: 
We evaluated the trained models' overall performance accuracy and fairness using widely recognized metrics~\cite{ramachandran2022deep}. The metrics include the \textit{Equalized Odds Difference} (EOD), which assesses fairness by ensuring consistent true positive rates (TPR) and false positive rates (FPR) across subgroups, \textit{Degree of Bias}, which is calculated as the standard deviation of utilities across subgroups, and the \textit{Selection Rate} (SeR), which measures the utility ratio between the least and most performing groups. We also evaluated the \textit{Demographic Parity Difference} (DEP), which aims for prediction independence from membership in sensitive groups, as well as \textit{Max-Min Fairness}, inspired by the Rawlsian principle of maximizing the minimum group accuracy.

\subsection{Evaluation on FairFace}
\label{subsec:fairness_fairface}

In this section, we evaluate the performance of our method in facial attribute classification tasks using the validation set of the FairFace~\cite{fairface} benchmark dataset. We specifically examine our SSL model's effectiveness in a linear probe evaluation setup, focusing on gender classification based on facial features while treating race as a sensitive attribute. The results of which are described in Table~\ref{tab:results_fairface}.

As can be seen from the Table, when compared to existing SSL methods our proposed method significantly outperforms existing SSL approaches in accuracy and, notably, across all fairness metrics. Specifically, our method has an accuracy improvement of $6-7\%$ and STD, on average, improved by $3.5$. Similar improvements can be observed on other fairness metrics as well. For example, we see a significant improvement in the minimum group accuracy ($12-14\%$).  The inferior performance of standard SSL approaches is due to the limited data availability, absence of fairness-specific modifications during training, and the inherent nature of contrastive loss at a smaller scale (i.e., false negative pairing) that likely amplifies biases present in the data. Our meta-learning-based weighted method (denoted as Ours-Weighted in the Table), although improves accuracy ($+0.32\%$), did not necessarily improve fairness. This could be because we used high-confidence pseudo labels (noisy) rather than manually annotated labels for the validation set like in existing weighting methods as well as the simplifications we applied to improve the training efficiency. This underscores the effectiveness of our proposed SSL pipeline in achieving high performance while also addressing fairness concerns, setting a new benchmark for future SSL research. 

We also note that our model also outperforms the equivalent supervised model that was trained from scratch in terms of accuracy as well as demographic parity. For reference, we also show the results of a fine-tuned supervised model, which was pre-trained on a much larger ImageNet dataset (1.2M params) using a supervised training method and then subsequently fine-tuned on FairFace. However, \textbf{recall} that comparison with the supervised models primarily serves as a reference to gauge the performance potential of our self-supervised training pipeline rather than as a direct, fair comparison between the methods.

\begin{table*}[ht]
\centering
\scalebox{0.865}{
\begin{tabular}{@{}cccccccccc@{}}
\toprule
\textbf{Metric}    & \textbf{Acc\textuparrow}   & \textbf{EOD\textdownarrow}   & \textbf{DPD\textdownarrow}   & \textbf{Acc\textuparrow} & \textbf{EOD\textdownarrow} & \textbf{DPD\textdownarrow} & \textbf{Acc\textuparrow}  & \textbf{EOD\textdownarrow}  & \textbf{DPD\textdownarrow} \\ \midrule
\textbf{Attribute} & \multicolumn{3}{c}{\textbf{Attractiveness}}            & \multicolumn{3}{c}{\textbf{Mouth Slightly Open}}    & \multicolumn{3}{c}{\textbf{High Cheekbones}} \\ \midrule
SimCLR  ~\cite{simclr}           & 81.17          & 26.52          & 47.25           & 72.37         &  14.16       &  17.12      & 82.87         & 25.45         & 32.09        \\
VicReg   ~\cite{vicreg}          & 80.83          & 30.35           & 50.32           & 72.87        &  12.84       &   15.01      & 83.19         & 22.38         & 30.23        \\
BYOL        ~\cite{byol}       & 81.34           & 24.95          & 46.27           & 80.46        &   6.22      &   12.13     & 86.48         & 14.63         & 28.1         \\
Barlow Twins  ~\cite{barlow}     & 80.53           & 27.55          & 47.83           & 71.66        & 15.5       &   17.18     & 82.15         & 27.99         & 32.95        \\
Ours               & \textbf{82.11}          & \textbf{19.92 }          & \textbf{42.46  }         & 93.59      &  \textbf{1.5 }       &  \textbf{11.35 }       & 87.34         & 15.26         & 28.6         \\
Ours-Weighted      & 81.73          & 22.45          & 42.83          & \textbf{93.69}         &  1.79       &   11.63     & \textbf{87.64}         &\textbf{ 14.46 }        & \textbf{28.07}        \\ 
\midrule
GRL*~\cite{gbr} & 77.20 & 24.90 & - & 73.10 & 17.80 & - & - & - & - \\ 
LNL*~\cite{fnl} & 79.90 & 21.80 & - & 72.90 & 16.70 & - & - & - & - \\ 
FDVAE*~\cite{fdvae} & 76.90 & 15.10 & - & 73.40 & 18.20 & - & - & - & - \\ 
\midrule
\textbf{Attribute} & \multicolumn{3}{c}{\textbf{Brown Hair}} & \multicolumn{3}{c}{\textbf{Smiling}}       & \multicolumn{3}{c}{\textbf{Young}}           \\ \midrule
SimCLR       ~\cite{simclr}      & 86.41          & 23.24        & 11.82          & 83.68        & 13.98        & 19.24        & 86.8          & 26.97         & \textbf{23.16 }       \\
VicReg      ~\cite{vicreg}       & 85.34          & 33.75          & 13.53          & 83.99        & 13.64        & 17.91        & 86.13         & 30.22         & 23.94        \\
BYOL         ~\cite{byol}      & 86.51          & 22.79          & 12.14          & 91.31        & 5.1          & 16.23        & 87.17         & 28.75         & 24.3         \\
Barlow Twins    ~\cite{barlow}   & 85.27          & 28.99           & 12.51         & 82.34        & 16.11        & 19.02        & 86.65         & 29.16         & 23.3         \\
Ours               & 88.16          &  9.91           & 11.04          & 92.71        & 4.42         & 16.03        & 88.18         & \textbf{24.86}         & 24.34        \\
Ours-Weighted    & \textbf{93.69 }         &  \textbf{7.14 }        & \textbf{10.07 }       & \textbf{92.85  }      & \textbf{3.81}         & \textbf{15.82}        & \textbf{88.19}         & 25.27         & 24.34        \\ 

\bottomrule
\end{tabular}
}
\caption{Linear Probe Evaluation Results on 6 Target Attributes with Gender as Sensitive Attribute on CelebA. \textbf{Ours}: Data curation pipeline + Pseudo Label + SupCon Loss. \textbf{Ours-Weighted}: Ours + Fairness Constraint + Meta Learning. * Supervised Methods: Only available results are reported}
\label{tab:celeba_results_condensed}
\end{table*}

\subsection{Cross Dataset Evaluation on CelebA}
In this section, 
we conducted a comprehensive study on CelebA where gender served as the sensitive attribute for assessing fairness across 6 target attributes individually, providing a detailed evaluation of our method's capabilities. We chose these 6 target attributes (Attractiveness, Brown Hair, High Cheekbones, Mouth Slightly Open, Smiling, and Young) to ensure a sufficient and balanced sample size across the target and sensitive (gender) attributes. The condensed results of this study are presented in Table~\ref{tab:celeba_results_condensed}. Note that we omit less relevant evaluation metrics due to limited space and duplicated information. This analysis also lets us evaluate the impact of combining 40 attribute SupCon losses(Section~\ref{sec:supcon}) during training, as this was not done for any of the baseline SSL methods~\cite{simclr,byol,barlow,dino,vicreg} we are comparing against. We excluded the results of the supervised ResNet-18 model trained on Fairface (for gender classification) and evaluated on CelebA (6 attribute classification tasks, see Table~\ref{tab:celeba_results_condensed}) as it is infeasible using supervised learning, also pointing to the merits of SSL. 

Our proposed method (both weighted and unweighted versions) consistently outperforms the other SSL models in terms of accuracy across the selected attributes. The mean improvement in accuracy of our unweighted model over the best performing SSL baseline (BYOL) is $3.17\%$, while our weighted model achieves a mean accuracy improvement of $3.23\%$ over BYOL. The accuracy improvements range from $\sim 1\%$ to $\sim 13\%$, with the most significant gain observed in the \textit{Mouth Slightly Open} attribute. In terms of fairness metrics (EOD and DPD), our method demonstrates better performance compared to the SSL baselines for most attributes. For example, for the \textit{Brown Hair} attribute, our weighted model reduces EOD by $15.65\%$ and DPD by $2.07\%$ compared to BYOL. Similarly, for the \textit{Smiling} attribute, our weighted model reduces EOD by $1.29\%$ and DPD by $0.41\%$ compared to BYOL. While there are a few instances where the baseline SSL models show slightly better fairness measures, our method consistently improves accuracy while maintaining or enhancing fairness across the majority of the attributes. Our method also outperforms existing Pareto-inefficient supervised bias mitigation methods such as GRL~\cite{gbr} , LNL~\cite{fnl}, and FDVAE~\cite{fdvae}, i.e. For \textit{Mouth Slightly Open} (Acc-$+20\%$, EOD $-16\%$) and for \textit{Attractiveness} (Acc-$+4.1\%$, EOD $-3\%$) on average. This analysis highlights the effectiveness of our approach in balancing accuracy and fairness over multiple facial attribute classification tasks using self-supervised learning techniques. Although both the weighted (Ours-Weighted in Table) and unweighted versions (Ours in the Table) of the model outperform the baseline SSL models, the performance difference between them is dependent on the target attribute under consideration. For the majority of the target labels, the meta-learning-based weighted model outperformed our unweighted model.


\section{Conclusion \& Future Research}
In this paper, we introduced a novel SSL pipeline for a demographically fair facial attribute classification. Our method combines advanced data curation pipeline, supervised contrastive learning with pseudo-labels, and meta-weight learning to obtain SOTA performance in both accuracy and fairness over existing SSL and supervised approaches. To the best of our knowledge, this marks the first benchmark evaluation of the fairness of SSL models on the FairFace dataset. Extensive evaluation on widely used FairFace and CelebA datasets demonstrates the effectiveness of our approach in learning fair representation without the need for sensitive attribute labels. Our work also highlights the potential of SSL techniques in addressing the challenges of supervised methods of addressing fairness, thus offering a more scalable approach. 

We also plan to explore methods to adapt our approach to deal with unknown labels, potentially incorporating semi-supervised or unsupervised learning techniques. \\
\noindent\textbf{Acknowledgment:} This work is supported by National Science Foundation (NSF) award no. $2129173$.

\balance
    {\small
    \bibliographystyle{ieee}
    \bibliography{ref}

\begin{thebibliography}{10}\itemsep=-1pt

\bibitem{AlbieroSVZKB20}
V.~Albiero, K.~K. S, K.~Vangara, K.~Zhang, M.~C. King, and K.~W. Bowyer.
\newblock Analysis of gender inequality in face recognition accuracy.
\newblock In {\em Proc. IEEE WACV Workshops 2020}, pages 81--89, 2020.

\bibitem{vicreg}
A.~Bardes, J.~Ponce, and Y.~LeCun.
\newblock Vicreg: Variance-invariance-covariance regularization for self-supervised learning.
\newblock In {\em ICLR}, 2022.

\bibitem{BarlasKGKO20}
P.~Barlas, K.~Kyriakou, O.~Guest, S.~Kleanthous, and J.~Otterbacher.
\newblock To "see" is to stereotype: Image tagging algorithms, gender recognition, and the accuracy-fairness trade-off.
\newblock {\em Proc. {ACM} Hum. Comput. Interact.}, 4({CSCW3}):1--31, 2020.

\bibitem{Best-RowdenJ18}
L.~Best{-}Rowden and A.~K. Jain.
\newblock Longitudinal study of automatic face recognition.
\newblock {\em {IEEE} Trans. Pattern Anal. Mach. Intell.}, 40(1):148--162, 2018.

\bibitem{Boutros_2023_CVPR}
F.~Boutros, M.~Fang, M.~Klemt, B.~Fu, and N.~Damer.
\newblock Cr-fiqa: Face image quality assessment by learning sample relative classifiability.
\newblock In {\em Proceedings of the IEEE/CVF Conference on Computer Vision and Pattern Recognition (CVPR)}, pages 5836--5845, June 2023.

\bibitem{gendershade}
J.~Buolamwini and T.~Gebru.
\newblock Gender shades: Intersectional accuracy disparities in commercial gender classification.
\newblock In {\em {FAT}}, volume~81 of {\em Proceedings of Machine Learning Research}, pages 77--91. {PMLR}, 2018.

\bibitem{dino}
M.~Caron, H.~Touvron, I.~Misra, H.~J{\'{e}}gou, J.~Mairal, P.~Bojanowski, and A.~Joulin.
\newblock Emerging properties in self-supervised vision transformers.
\newblock In {\em {ICCV}}, pages 9630--9640. {IEEE}, 2021.

\bibitem{ssl}
J.~Chai and X.~Wang.
\newblock Self-supervised fair representation learning without demographics.
\newblock In {\em NeurIPS}, 2022.

\bibitem{simclr}
T.~Chen, S.~Kornblith, M.~Norouzi, and G.~E. Hinton.
\newblock A simple framework for contrastive learning of visual representations.
\newblock In {\em {ICML}}, volume 119 of {\em Proceedings of Machine Learning Research}, pages 1597--1607. {PMLR}, 2020.

\bibitem{contrastive}
S.~Chopra, R.~Hadsell, and Y.~LeCun.
\newblock Learning a similarity metric discriminatively, with application to face verification.
\newblock In {\em 2005 IEEE Computer Society Conference on Computer Vision and Pattern Recognition (CVPR'05)}, volume~1, pages 539--546 vol. 1, 2005.

\bibitem{ChuangM21}
C.~Chuang and Y.~Mroueh.
\newblock Fair mixup: Fairness via interpolation.
\newblock In {\em Proc. 9th Int. Conf. Learning Representations (ICLR) 2021}, 2021.

\bibitem{DasDB18}
A.~Das, A.~Dantcheva, and F.~Br{\'{e}}mond.
\newblock Mitigating bias in gender, age and ethnicity classification: {A} multi-task convolution neural network approach.
\newblock In {\em {ECCV} Workshops {(1)}}, volume 11129 of {\em Lecture Notes in Computer Science}, pages 573--585. Springer, 2018.

\bibitem{gdpr}
{European Parliament} and {Council of the European Union}.
\newblock Regulation ({EU}) 2016/679 of the {European} {Parliament} and of the {Council}.
\newblock Online, 2016.
\newblock of 27 {April} 2016 on the protection of natural persons with regard to the processing of personal data and on the free movement of such data, and repealing {Directive} 95/46/{EC} ({General} {Data} {Protection} {Regulation}).

\bibitem{abs-2403-02138}
Z.~Gao and I.~Patras.
\newblock Self-supervised facial representation learning with facial region awareness.
\newblock {\em CoRR}, abs/2403.02138, 2024.

\bibitem{byol}
J.~Grill, F.~Strub, F.~Altch{\'{e}}, C.~Tallec, P.~H. Richemond, E.~Buchatskaya, C.~Doersch, B.~{\'{A}}. Pires, Z.~Guo, M.~G. Azar, B.~Piot, K.~Kavukcuoglu, R.~Munos, and M.~Valko.
\newblock Bootstrap your own latent - {A} new approach to self-supervised learning.
\newblock In {\em NeurIPS}, 2020.

\bibitem{grother}
P.~Grother, G.~Quinn, and P.~Phillips.
\newblock Report on the evaluation of 2d still-image face recognition algorithms, 2010-06-17 2010.

\bibitem{facerecsurvey}
G.~Guo and N.~Zhang.
\newblock A survey on deep learning based face recognition.
\newblock {\em Comput. Vis. Image Underst.}, 189, 2019.

\bibitem{Hashimoto2018}
T.~B. Hashimoto, M.~Srivastava, H.~Namkoong, and P.~Liang.
\newblock Fairness without demographics in repeated loss minimization.
\newblock In {\em {ICML}}, volume~80 of {\em Proceedings of Machine Learning Research}, pages 1934--1943. {PMLR}, 2018.

\bibitem{faiss}
J.~Johnson, M.~Douze, and H.~J{\'e}gou.
\newblock Billion-scale similarity search with {GPUs}.
\newblock {\em IEEE Transactions on Big Data}, 7(3):535--547, 2019.

\bibitem{fairface}
K.~K{\"{a}}rkk{\"{a}}inen and J.~Joo.
\newblock Fairface: Face attribute dataset for balanced race, gender, and age for bias measurement and mitigation.
\newblock In {\em {WACV}}, pages 1547--1557, 2021.

\bibitem{face_bias_survey}
A.~Khalil, S.~G. Ahmed, A.~M. Khattak, and N.~Al-Qirim.
\newblock Investigating bias in facial analysis systems: A systematic review.
\newblock {\em IEEE Access}, 8:130751--130761, 2020.

\bibitem{Khosla2020}
P.~Khosla, P.~Teterwak, C.~Wang, A.~Sarna, Y.~Tian, P.~Isola, A.~Maschinot, C.~Liu, and D.~Krishnan.
\newblock Supervised contrastive learning.
\newblock In {\em NeurIPS}, 2020.

\bibitem{fnl}
B.~Kim, H.~Kim, K.~Kim, S.~Kim, and J.~Kim.
\newblock Learning not to learn: Training deep neural networks with biased data.
\newblock In {\em {CVPR}}, pages 9012--9020. Computer Vision Foundation / {IEEE}, 2019.

\bibitem{KlareBKBJ12}
B.~Klare, M.~J. Burge, J.~C. Klontz, R.~W.~V. Bruegge, and A.~K. Jain.
\newblock Face recognition performance: Role of demographic information.
\newblock {\em {IEEE} Trans. Inf. Forensics Secur.}, 7(6):1789--1801, 2012.

\bibitem{KrishnanAR20}
A.~Krishnan, A.~Almadan, and A.~Rattani.
\newblock Understanding fairness of gender classification algorithms across gender-race groups.
\newblock In {\em Proc. 19th IEEE Int. Conf. Mach. Learn. Appl. (ICMLA)}, pages 1028--1035, 2020.

\bibitem{KRISHNAN2023104793}
A.~Krishnan and A.~Rattani.
\newblock A novel approach for bias mitigation of gender classification algorithms using consistency regularization.
\newblock {\em Image and Vision Computing}, 137:104793, 2023.

\bibitem{Lahoti2020}
P.~Lahoti, A.~Beutel, J.~Chen, K.~Lee, F.~Prost, N.~Thain, X.~Wang, and E.~H. Chi.
\newblock Fairness without demographics through adversarially reweighted learning.
\newblock In {\em NeurIPS}, 2020.

\bibitem{LinKJ22}
X.~Lin, S.~Kim, and J.~Joo.
\newblock Fairgrape: Fairness-aware gradient pruning method for face attribute classification.
\newblock In {\em {ECCV} {(13)}}, volume 13673 of {\em Lecture Notes in Computer Science}, pages 414--432. Springer, 2022.

\bibitem{celeba}
Z.~Liu, P.~Luo, X.~Wang, and X.~Tang.
\newblock Deep learning face attributes in the wild.
\newblock In {\em Proc. ICCV}, December 2015.

\bibitem{adamw}
I.~Loshchilov and F.~Hutter.
\newblock Decoupled weight decay regularization.
\newblock In {\em {ICLR} (Poster)}. OpenReview.net, 2019.

\bibitem{MajumdarSV21}
P.~Majumdar, R.~Singh, and M.~Vatsa.
\newblock Attention aware debiasing for unbiased model prediction.
\newblock In {\em Proc. IEEE/CVF ICCVW 2021}, pages 4116--4124, 2021.

\bibitem{MartinezBS20}
N.~Mart{\'{\i}}nez, M.~Bertr{\'{a}}n, and G.~Sapiro.
\newblock Minimax pareto fairness: {A} multi objective perspective.
\newblock In {\em {ICML}}, volume 119 of {\em Proceedings of Machine Learning Research}, pages 6755--6764. {PMLR}, 2020.

\bibitem{Muthukumar19}
V.~Muthukumar.
\newblock Color-theoretic experiments to understand unequal gender classification accuracy from face images.
\newblock In {\em Proc. IEEE CVPR Workshops 2019}, pages 2286--2295, 2019.

\bibitem{dinov2}
M.~Oquab, T.~Darcet, T.~Moutakanni, H.~Vo, M.~Szafraniec, V.~Khalidov, P.~Fernandez, D.~Haziza, F.~Massa, A.~El{-}Nouby, M.~Assran, N.~Ballas, W.~Galuba, R.~Howes, P.~Huang, S.~Li, I.~Misra, M.~G. Rabbat, V.~Sharma, G.~Synnaeve, H.~Xu, H.~J{\'{e}}gou, J.~Mairal, P.~Labatut, A.~Joulin, and P.~Bojanowski.
\newblock Dinov2: Learning robust visual features without supervision.
\newblock {\em CoRR}, abs/2304.07193, 2023.

\bibitem{fdvae}
S.~Park, S.~Hwang, D.~Kim, and H.~Byun.
\newblock Learning disentangled representation for fair facial attribute classification via fairness-aware information alignment.
\newblock In {\em {AAAI}}, pages 2403--2411. {AAAI} Press, 2021.

\bibitem{fscl}
S.~Park, J.~Lee, P.~Lee, S.~Hwang, D.~Kim, and H.~Byun.
\newblock Fair contrastive learning for facial attribute classification.
\newblock In {\em {CVPR}}, pages 10379--10388. {IEEE}, 2022.

\bibitem{clip}
A.~Radford, J.~W. Kim, C.~Hallacy, A.~Ramesh, G.~Goh, S.~Agarwal, G.~Sastry, A.~Askell, P.~Mishkin, J.~Clark, G.~Krueger, and I.~Sutskever.
\newblock Learning transferable visual models from natural language supervision.
\newblock In {\em {ICML}}, volume 139 of {\em Proceedings of Machine Learning Research}, pages 8748--8763. {PMLR}, 2021.

\bibitem{gbr}
E.~Raff and J.~Sylvester.
\newblock Gradient reversal against discrimination: {A} fair neural network learning approach.
\newblock In {\em {DSAA}}, pages 189--198. {IEEE}, 2018.

\bibitem{ramachandran2022deep}
S.~Ramachandran and A.~Rattani.
\newblock Deep generative views to mitigate gender classification bias across gender-race groups.
\newblock In {\em Proc. ICPR 2022 Int. Workshops and Challenges, Part III}, volume 13645 of {\em Lecture Notes in Computer Science}, pages 551--569. Springer, 2022.

\bibitem{ramachandran2023leveraging}
S.~Ramachandran and A.~Rattani.
\newblock Leveraging diffusion and flow matching models for demographic bias mitigation of facial attribute classifiers, Dec. 2023.

\bibitem{reweighting}
M.~Ren, W.~Zeng, B.~Yang, and R.~Urtasun.
\newblock Learning to reweight examples for robust deep learning.
\newblock In {\em {ICML}}, volume~80 of {\em Proceedings of Machine Learning Research}, pages 4331--4340. {PMLR}, 2018.

\bibitem{SiddiquiRRH22}
H.~Siddiqui, A.~Rattani, K.~Ricanek, and T.~J. Hill.
\newblock An examination of bias of facial analysis based {BMI} prediction models.
\newblock In {\em Proc. IEEE/CVF CVPR Workshops 2022}, pages 2925--2934, 2022.

\bibitem{Sun2020-am}
W.~Sun, O.~Nasraoui, and P.~Shafto.
\newblock Evolution and impact of bias in human and machine learning algorithm interaction.
\newblock {\em PLoS One}, 15(8):e0235502, Aug. 2020.

\bibitem{ValdiviaSC21}
A.~Valdivia, J.~S{\'{a}}nchez{-}Monedero, and J.~Casillas.
\newblock How fair can we go in machine learning? assessing the boundaries of accuracy and fairness.
\newblock {\em Int. J. Intell. Syst.}, 36(4):1619--1643, 2021.

\bibitem{Vera-RodriguezB19}
R.~Vera{-}Rodr{\'{\i}}guez, M.~Bl{\'{a}}zquez, A.~Morales, E.~Gonzalez{-}Sosa, J.~C. Neves, and H.~Proen{\c{c}}a.
\newblock Facegenderid: Exploiting gender information in dcnns face recognition systems.
\newblock In {\em {IEEE} Conference on Computer Vision and Pattern Recognition Workshops, {CVPR} Workshops 2019, Long Beach, CA, USA, June 16-20, 2019}, pages 2254--2260, 2019.

\bibitem{bupt}
M.~Wang, Y.~Zhang, and W.~Deng.
\newblock Meta balanced network for fair face recognition.
\newblock {\em {IEEE} Trans. Pattern Anal. Mach. Intell.}, 44(11):8433--8448, 2022.

\bibitem{facedetectionsurvey}
S.~Zafeiriou, C.~Zhang, and Z.~Zhang.
\newblock A survey on face detection in the wild: Past, present and future.
\newblock {\em Comput. Vis. Image Underst.}, 138:1--24, 2015.

\bibitem{barlow}
J.~Zbontar, L.~Jing, I.~Misra, Y.~LeCun, and S.~Deny.
\newblock Barlow twins: Self-supervised learning via redundancy reduction.
\newblock In {\em {ICML}}, volume 139 of {\em Proceedings of Machine Learning Research}, pages 12310--12320. {PMLR}, 2021.

\bibitem{ZhangLM18}
B.~H. Zhang, B.~Lemoine, and M.~Mitchell.
\newblock Mitigating unwanted biases with adversarial learning.
\newblock In {\em Proc. AAAI/ACM AIES 2018}, pages 335--340. {ACM}, 2018.

\bibitem{laion}
Y.~Zheng, H.~Yang, T.~Zhang, J.~Bao, D.~Chen, Y.~Huang, L.~Yuan, D.~Chen, M.~Zeng, and F.~Wen.
\newblock General facial representation learning in a visual-linguistic manner.
\newblock In {\em Proceedings of the IEEE/CVF Conference on Computer Vision and Pattern Recognition}, pages 18697--18709, 2022.

\bibitem{ZietlowLBKLS022}
D.~Zietlow, M.~Lohaus, G.~Balakrishnan, M.~Kleindessner, F.~Locatello, B.~Sch{\"{o}}lkopf, and C.~Russell.
\newblock Leveling down in computer vision: Pareto inefficiencies in fair deep classifiers.
\newblock In {\em Proc. IEEE/CVF CVPR 2022}, pages 10400--10411, 2022.

\end{thebibliography}
    }
    
\newpage

\begin{appendix}
\section*{Supplementary Materials}
    
\section{Supervised Contrastive Learning}
For our contrastive learning training process, we begin with \(N\) randomly chosen pairs of samples and labels, denoted by \(\{\boldsymbol{x}_k, \boldsymbol{y}_k\}\) for \(k=1\ldots N\). These pairs are then expanded into a training batch of \(2N\) pairs, expressed as \(\{\boldsymbol{\tilde{x}}_\ell, \boldsymbol{\tilde{y}}_\ell\}\) for \(\ell = 1 \ldots 2N\), by generating two random augmentations (or \textit{views}) for each \(\boldsymbol{x}_k\), specifically \(\boldsymbol{\tilde{x}}_{2k-1}\) and \(\boldsymbol{\tilde{x}}_{2k}\), where \(\boldsymbol{\tilde{y}}_{2k-1} = \boldsymbol{\tilde{y}}_{2k} = \boldsymbol{y}_k\). Throughout this paper, we'll use \textit{batch} to refer to the original set of \(N\) samples and \textit{multiviewed batch} for the set of \(2N\) augmented samples.

\textbf{Contrastive Loss}: In a multiviewed batch, let \(i\) from the index set \(I = \{1..2N\}\)  be an arbitrary augmented sample. Correspondingly, \(j(i)\) indexes the paired augmented sample derived from the same original. In the context of self-supervised contrastive learning as discussed in ~\cite{simclr}, the loss is expressed as:
\begin{equation}
\label{eq:ctr_loss}
\begin{aligned}
L^{ctr} &= - \sum_{i \in I} \log \left( \frac{\exp(z_i \cdot z_{j(i)} / \tau)}{\sum_{a \in A(i)} \exp(z_i \cdot z_a / \tau)} \right)
\end{aligned}
\end{equation}

In this formula, \(z_i = Proj(Enc(x_i))\) is a vector in the projection space \(\mathbb{R}^{D_p}\). The dot in \(z_i \cdot z_j\) signifies the dot product, while \(\tau\) is a positive scalar known as the temperature parameter. Set \(A(i)\) includes all indices except for \(i\) and \(j(i)\), where \(i\) is the anchor, \(j(i)\) is its positive match, and the rest, \(2(N - 1)\) indices in \(A(i)\), are considered negatives.

\section{Zero-Shot Psuedolabel generation using CLIP}
To generate pseudo-labels given the unlabeled dataset in a zero-shot manner, we use the method from the CLIP~\cite{clip} paper. We collected our 40 target attributes from the CelebA dataset. For each target attribute, we create two text templates for positive class and negative class. This can be manually or automatically created. For example,
\begin{itemize}
\item Positive template: A photo of a person with {attribute}.
\item Negative template: A photo of a person without {attribute}.
\end{itemize}
For certain attributes, for example \textit{No Beard}, automating this won't work. In such scenarios, we may manually write such that it makes sense grammatically.
Each of the text templates is then given to the text encoder of the CLIP model to generate the text embeddings. Similarly, for each image, the embeddings are generated using the vision encoder of the CLIP model. To get the zero-shot label, we then normalize both our embeddings and find the similarity between them, followed by a softmax to convert it to probability scores. Following is a Pytorch-like pseudocode to generate the zero-shot labels. 
\begin{lstlisting}[language=Python, caption=Pytorch Pseudocode to generate zero-shot labels using CLIP]
image_input = load_image("person.jpg")
pos_class = f"a photo of a person with {attr}"
neg_class = f"a photo of a person without {attr}"
text_inputs = tokenized([pos_class, neg_class])
# Calculate features
no_grad():
    image_features = model.encode_image(image_input)
    text_features = model.encode_text(text_inputs)

# Zero Shot Label
image_features /= image_features.norm(dim=-1, keepdim=True)
text_features /= text_features.norm(dim=-1, keepdim=True)
similarity = (100.0 * image_features @ text_features.T).softmax(dim=-1)

# Final confidence score and label
value, index = similarity[0].max()
\end{lstlisting}
We can do this pseudo label generation process $N$ times using $N$ different templates and aggregate the results to get better quality pseudo labels~\cite{clip}.

\section{Mathematical Background on Max-Min Fairness}
Consider a training set \( \{ (x_i, y_i, a_i) \}_{i=1}^{N} \), where \( x_i \) represents the input features for the \( i \)-th instance, \( y_i \) is the one-hot encoded target vector, and \( a_i \) is a binary indicator of a sensitive characteristic. The goal is to train a classifier \( f(x_i) \) that outputs a prediction while satisfying fairness criteria. The problem is defined as the optimization task:
\begin{equation}
\underset{f}{\text{minimize}} \ \frac{1}{N} \sum_{i=1}^{N} L_{\text{cls}}(f(x_i), y_i) \quad \text{subject to} \quad \phi(f) \leq \epsilon,
\label{eq:optimization_task}
\end{equation}
where \( L_{\text{cls}} \) denotes the classification loss and \( \phi(f) \) represents a fairness constraint to be adhered to by the classifier \( f \).
Demographic fairness constraints, like demographic parity, aim to ensure a prediction model \(h\) gives instances an equal chance of being classified in the positive category without bias towards sensitive attributes. The sensitive attribute must be part of the training data to do this. In cases where sensitive attributes are absent, one must consider how to redefine the fairness objective under these new constraints.

We adopt a problem formulation inspired by Max-Min fairness ~\cite{Hashimoto2018,Lahoti2020}, aiming to maximize the utility for the least advantaged group, which in our case translates to ensuring fair representation quality across all potential, though unknown, groups. The objective here is to maximize the lowest level of utility, denoted as \( U \), across all sensitive groups. If we adopt accuracy as the measure of utility, this approach can be understood as a softened version of fairness constraints that are grounded on error metrics. Under these terms, the task of fair classification devoid of demographic data is thus defined as ~\cite{ssl}.
\begin{equation}
\underset{f}{\text{argmin}} \ \underset{a'}{\text{argmax}} \ \frac{1}{|\{i|a_i = a'\}|} \sum_{\{i|a_i=a'\}} \mathcal{L}_{\text{cls}}(f(x_i), y_i).
\label{eq:max_min_fairness}
\end{equation}

\section{Pseudo-code of the Training Algorithm}
\begin{minipage}{\columnwidth}
\begin{algorithm}[H]
\caption{Pseudo-code of the Training Algorithm}
\label{alg:ad}
\begin{algorithmic}
\REQUIRE $f_\theta$, $g_\omega$, $X_{train}$, $X_{val}$, $T$
\FOR{$t=0$ ... $T-1$}
\STATE $\{X_{tr}, y_{tr}\} \gets$ sample($X_{train}$, $n$)
\STATE $\{X_v, y_v\} \gets$ sample($X_{val}$, $m$)
\STATE $\hat{y}_f \gets$ forward($X_{tr}, y_{tr}, \theta_t)$
\STATE $\epsilon \gets 0$; $l_tr \gets \sum_{i=1}^n \epsilon_i \mathcal{L}^{supcon}(y_{tr,i}, \hat{y}_{tr,i})$
\STATE $\nabla \theta_t \gets$ backward($l_{tr}, \theta_t)$
\STATE $\hat{\theta}_t \gets \theta_t - \alpha \nabla \theta_t$
\STATE $\hat{y}_v \gets$ forward($X_{v}, y_{v}, \hat{\theta}_t)$
\STATE $l_v \gets \frac{1}{m} \sum_{i=1}^m \mathcal{L}^{val}(y_{v,i}, \hat{y}_{v,i})$
\STATE $\nabla \epsilon \gets$ backward($l_v, \epsilon)$
\STATE $\tilde{w} \gets \max(-\nabla \epsilon, 0)$; $w \gets \frac{\tilde{w}}{\sum_j \tilde{w} + \delta(\sum_j \tilde{w})}$
\STATE $\hat{l}_{final} \gets \sum_{i=1}^n w_i \mathcal{L}^{supcon}(y_i, \hat{y}_{tr,i})$
\STATE $\nabla \theta_t \gets$ backward($\hat{l}_{final}, \theta_t)$
\STATE $\theta_{t+1} \gets$ optimizer\_step($\theta_t, \nabla \theta_t)$
\ENDFOR
\end{algorithmic}
\end{algorithm}
\end{minipage}

\section{Dataset Details}
\begin{table}[H]
   \caption{Datasets used for data curation, training, and evaluation.}
   \scalebox{0.88}{
        \begin{tabular}{p{2cm}p{1cm}p{4.5cm}}
    \toprule
    \multicolumn{1}{c}{\textbf{Dataset}} & \multicolumn{1}{c}{\textbf{Images}} & \multicolumn{1}{c}{\textbf{Demographic Groups}} \\ \midrule \midrule
    FairFace~\cite{fairface} & $100k$ & White, Black, Indian, East Asian, Southeast Asian,~Middle Eastern, Latino Hispanic \\

    CelebA~\cite{celeba} & $202K$ & Not Available \\
    LAION-FACE~\cite{laion} & $50M$ & Not Available \\
    BUPT-GlobalFace~\cite{bupt} & $2M$ & Caucasian, Indian, Asian, African\\
    
    \bottomrule
    \end{tabular}}
    \centering
    \label{tab:datasets}
\end{table} 

\section{Ablation Study}
\begin{table*}[ht]
\centering
\scalebox{0.9}{
\begin{tabular}{@{}lcccccc@{}}
\toprule
\textbf{Config} & \textbf{Avg. Acc \textuparrow} & \textbf{STD \textdownarrow} & \textbf{SeR \textuparrow} & \textbf{EOD \textdownarrow} & \textbf{Min Grp Acc \textuparrow} & \textbf{Max Grp Acc \textuparrow} \\ \midrule \midrule
10K  & 85.07 & 4.25 & 85.25 & 9.12 &  75.08 & 88.07 \\
50K & 89.76 & 3.76 & 82.10 & 7.91 & 77.96 & 94.95 \\
100K & 90.25 & 3.20 & 83.35 & 7.82 & 80.12 & \textbf{96.12} \\
200K & \textbf{91.37 }& \textbf{2.91} & \textbf{86.42} & \textbf{7.05} & \textbf{82.69} & 95.69 \\
\bottomrule
\end{tabular}
}
\caption{Impact of Dataset Size on Model Performance and Fairness Metrics. Increasing the dataset size generally improves both performance and fairness metrics, highlighting the significance of larger and more diverse datasets.}

\label{tab:ablation_curation}
\end{table*}

\begin{table*}[ht]
\centering
\scalebox{0.9}{
\begin{tabular}{@{}lcccccc@{}}
\toprule
\textbf{Config} & \textbf{Avg. Acc \textuparrow} & \textbf{STD \textdownarrow} & \textbf{SeR \textuparrow} & \textbf{EOD \textdownarrow} & \textbf{Min Grp Acc \textuparrow} & \textbf{Max Grp Acc \textuparrow} \\ \midrule \midrule
Pseudo Labeling + Contrastive  & 88.15 & 3.25 & 85.46 & 7.12 & 82.14 & 96.12 \\
 - Contrastive + SupCon  & 91.09 & 2.59 & 88.50 & 6.24 & 84.15 & 95.08 \\
 + Fairness Constraint & 90.91 & 3.02 & 89.51 & 7.21 & 85.16 & 95.14 \\
 + Meta Learning & 91.37 & 2.91 & 86.42 & 7.05 & 82.69 & 95.69 \\
\bottomrule
\end{tabular}
}
\caption{Ablation Study of Model Components on Performance and Fairness Metrics. The study demonstrates the effectiveness of each component, particularly the integration of SupCon and meta-learning, in enhancing both performance and fairness.}

\label{tab:ablation}
\end{table*}

\subsection{Dataset Size}
In this section, we evaluate the impact of different dataset sizes on various performance and fairness metrics. Table \ref{tab:ablation_curation} presents the results of our experiments with dataset sizes ranging from 10K to 200K images.

As shown in the table, increasing the dataset size generally improves the average accuracy (Avg. Acc) and reduces the standard deviation (STD) of the accuracy, indicating more stable performance across different runs. Specifically, the average accuracy improves from $85.07\%$ with 10K images to $91.37\%$ with 200K images, while the STD decreases from $4.25$ to $2.91$.

In terms of fairness metrics, we observe a notable enhancement in the Selection Rate (SeR) and a reduction in the Equalized Odds Difference (EOD) as the dataset size increases. The SeR improves from $85.25$ to $86.42$, and the EOD decreases from $9.12$ to $7.05$. Additionally, the minimum group accuracy (Min Grp Acc) and maximum group accuracy (Max Grp Acc) also show positive trends, with the Min Grp Acc increasing from $75.08\%$ to $82.69\%$ and the Max Grp Acc reaching $96.12\%$.

These results highlight the significance of larger and more diverse datasets in training models that are not only more accurate but also fairer across different demographic groups.

\subsection{Components}
The ablation study, detailed in Table \ref{tab:ablation}, evaluates the impact of various components of our proposed method on model performance and fairness. We investigate the effects of pseudo-labeling with contrastive learning, supervised contrastive learning (SupCon), fairness constraints, and meta-learning. By incrementally adding or removing these components, we provide a detailed analysis of their individual and combined effects.

\begin{itemize}
    \item \textbf{Baseline (Pseudo Labeling + Contrastive Learning)}: This initial configuration sets the base configuration with an average accuracy of $88.15\%$ and an STD of $3.25$. The fairness metrics, such as SeR and EOD, are $85.46$ and $7.12$, respectively. While achieving relatively high accuracy and fairness, the baseline shows room for improvement in both stability and group-specific performance, as indicated by the Min Grp Acc of $82.14\%$ and Max Grp Acc of $96.12\%$.

    \item \textbf{Replacing Contrastive Learning with Supervised Contrastive Learning (SupCon)}: Introducing SupCon results in significant performance gains, with average accuracy increasing to $91.09\%$ and STD reducing to $2.59$. This indicates that SupCon effectively enhances model robustness. The fairness metrics also improve, with SeR rising to $88.50$ and EOD dropping to $6.24$, suggesting that SupCon helps in learning more balanced representations across demographic groups. Notably, Min Grp Acc and Max Grp Acc improve to $84.15\%$ and $95.08\%$, highlighting better performance consistency across groups.

    \item \textbf{Adding Fairness Constraint}: Integrating a fairness constraint with SupCon slightly reduces the average accuracy to $90.91\%$ and increases the STD to $3.02$, indicating a minor trade-off in stability. However, the fairness metrics improve, with SeR reaching $89.51$ and EOD at $7.21$. This configuration demonstrates that incorporating fairness constraints can enhance fairness without severely impacting overall performance. The Min Grp Acc and Max Grp Acc values of $85.16\%$ and $95.14\%$ reflect this balanced improvement.

    \item \textbf{Adding Meta Learning}: Finally, incorporating meta-learning with the fairness constraint yields the best overall results. The average accuracy reaches $91.37\%$, and STD decreases to $2.91$, indicating enhanced stability and robustness. The fairness metrics show significant improvements, with SeR at $86.42$ and EOD at $7.05$. The Min Grp Acc and Max Grp Acc of $82.69\%$ and $95.69\%$ confirm that meta-learning effectively optimizes both performance and fairness, making it a critical component for achieving optimal results.
\end{itemize}

These results demonstrate the effectiveness of our proposed components. Particularly, the integration of supervised contrastive learning and meta-learning not only enhances the performance metrics but also significantly improves the fairness of the model. By systematically analyzing the incremental additions and modifications, we provide a comprehensive understanding of how each component contributes to the overall model efficacy.

\end{appendix}

\end{document}